\pdfoutput=1
\documentclass[11pt]{scaleai-paper}

\usepackage{amsmath}
\usepackage{amsfonts}
\usepackage{amssymb}
\usepackage{amsthm}
\usepackage{booktabs}
\usepackage{tabularx}
\usepackage{tabulary}
\usepackage{multirow}
\usepackage{subcaption}
\usepackage{float}
\usepackage[square,numbers,sort&compress]{natbib}
\usepackage{xspace}
\usepackage{url}
\usepackage[colorlinks=true,linkcolor=scaleLink,citecolor=scaleLink,urlcolor=scaleLink]{hyperref}
\usepackage[capitalise,nameinlink]{cleveref}

\let\svthefootnote\thefootnote
\newcommand\freefootnote[1]{%
  \let\thefootnote\relax%
  \footnotetext{#1}%
  \let\thefootnote\svthefootnote%
}

\newtcolorbox{appendixprompt}{
  enhanced,
  breakable,
  colback=black!4,
  colframe=black!40,
  boxrule=0.4pt,
  arc=2pt,
  outer arc=2pt,
  boxsep=0pt,
  left=8pt,
  right=8pt,
  top=7pt,
  bottom=7pt,
  before skip=10pt,
  after skip=10pt,
  fontupper=\normalsize\normalfont\rmfamily,
  before upper={\setlength{\parindent}{0pt}\setlength{\parskip}{0pt}}
}

\newcommand{\benchname}{\textsc{SG-LegalClarify}\xspace}

\papertype{Scale AI Paper Template}
\contact{\texttt{udari.sehwag@scale.com, prep@scale.com} \quad | \quad \url{https://labs.scale.com/papers/sglegalclarify}}

\title{SG-LegalClarify: A Large-Scale Benchmark for Legal Reasoning and Ambiguity Recognition in Singapore Law}

\author[1, 4*]{Ziwei Gu}
\author[2*$\dagger$]{Michael S. Lee}
\author[3]{Drew Rein}
\author[2]{Yash Maurya}
\author[5]{Vikneswaran Kumaran}
\author[5]{Darshini Ramiah}
\author[6]{Chenguang Wang}
\author[2]{Udari Madhushani Sehwag}

\affil[1]{Human Frontier Collective, Scale AI}
\affil[2]{Preparedness and Policy, Scale AI}
\affil[3]{Scale AI}
\affil[4]{Harvard University}
\affil[5]{Infocomm Media Development Authority (IMDA)}
\affil[6]{University of California, Santa Cruz}

\begin{document}

\freefootnote{\textsuperscript{*}Equal contribution.}
\freefootnote{${}^\dagger$Work conducted while at Scale AI.}

\maketitle

\begin{abstract}
Singapore offers a distinctive testbed for legal reasoning by large language models (LLMs). Its legal system combines inherited English common law with jurisdiction-specific statutes, requiring models to distinguish local doctrine from generic common-law intuition. This setting also highlights the challenge of recognizing underspecification in lay legal queries, because the facts material to a legal outcome depend on the governing doctrine. Yet legal benchmarks rarely cover Singapore law or test this form of ambiguity recognition. To address these gaps, we introduce \benchname, a large-scale benchmark for open-ended legal reasoning and ambiguity recognition in Singapore law. \benchname contains 1{,}100 lawyer-authored source questions spanning 10 legal domains, including 340 fully specified and 760 underspecified questions. Each underspecified question has a fixed hidden-fact assignment and a fully specified counterpart, yielding 1{,}860 evaluation instances. Responses are graded with lawyer-authored Issue, Rule, Application, and Conclusion (IRAC) rubrics. We evaluate 19 LLMs, closed-book and with web search, using a multi-turn pipeline in which models decide whether to answer or clarify, and a user simulator reveals hidden facts only when asked. We find the benchmark far from saturated: the best models satisfy only 69.8\% of mandatory rubric items closed-book (average 53.8\%) and 77.8\% with web search (average 63.0\%). Models also struggle to recognize when clarification is necessary. 
We release the public split of our benchmark\footnote{\url{https://huggingface.co/datasets/ScaleAI/SG-LegalClarify}}, along with the model outputs on it and the evaluation pipeline\footnote{\url{https://github.com/scaleapi/SG-LegalClarify}}, to support future research.
\end{abstract}

\section{Introduction}

Singapore offers a distinctive setting for evaluating legal AI. Its legal system is among the most highly regarded in the world, ranking in the global top ten on civil and criminal justice~\citep{wjp2025index}. It is also a common-law jurisdiction in which inherited English law applies but subject to local modification and Singapore-specific statutes~\citep{aela1993}. A model therefore cannot rely on generic common-law intuition only and must identify when Singapore-specific doctrine governs.

\begin{figure}[t]
  \centering
  \includegraphics[width=\textwidth]{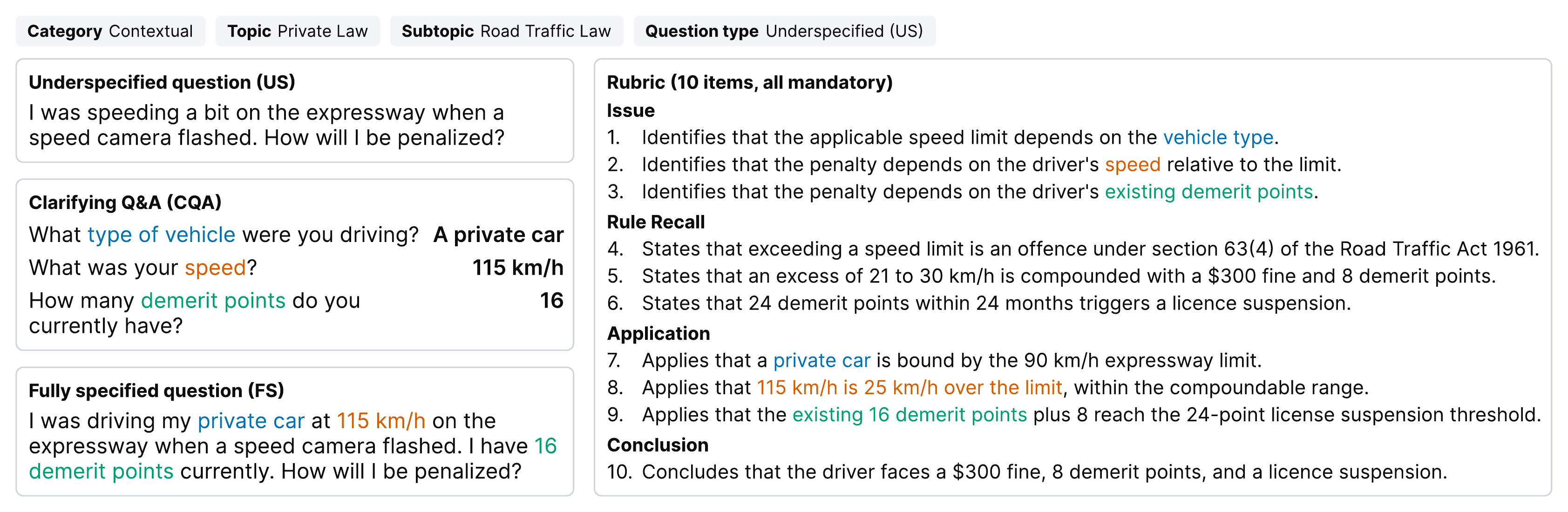}
  \caption{An example source question from \benchname. The underspecified (US) question omits three legally material facts, the clarifying questions and answers (CQA) reveal them, and the fully specified (FS) version states them outright. Responses are graded against a lawyer-written IRAC rubric. Colors mark the same variable across CQA, FS, and rubric.}
  \label{fig:example}
\end{figure}

The difficulty is compounded when large language models (LLMs) give legal guidance to lay users, who often do not know which facts the law makes material~\citep{cheong2024lawyer,kapoor2024promises}, resulting in underspecified queries (Figure~\ref{fig:example}). To answer such queries, a model must clarify selectively, asking when a material fact is missing even if the query is linguistically clear, and answering directly when the information is sufficient, rather than burdening users with unnecessary questions. The clarification decision is itself part of legal reasoning, since recognizing that a fact is missing requires identifying the legal issue and recalling the rule that makes the fact material.

Existing legal benchmarks leave gaps in assessing selective clarification and do not pose clarification as a decision. Most evaluate models on fixed inputs treated as answerable as given, emphasizing legal knowledge and reasoning but not whether a model recognizes that legally material information is missing and seeks clarification~\citep{guha2023legalbench,fan2026lexam,fei2024lawbench,li2024lexeval}. Recent consultation benchmarks do require models to elicit facts through multi-turn dialogue~\citep{shi2026plawbench,yuan2026lecode,bystronski2026factual,zhang2026dlawbench}, but they focus on queries requiring clarification only, so they measure whether a model clarifies, not whether it clarifies only when it should. Additionally, failures can also be harder to trace, as a flawed conclusion may stem from a key fact the model fails to elicit, or a misidentified issue or  rule~\citep{dahl2024large}. Rubric-based benchmarks are beginning to localize which criteria a response misses~\citep{chlapanis2025greekbarbench,lee2026evaluating,shi2026plawbench,zhang2026dlawbench}, but selective clarification requires evaluation framework beyond fine-grained rubrics. For Singapore, existing work focuses on compliance detection, embedding retrieval, or citation retrieval~\citep{nguyen2025autolaw,butler2025mleb,ern2026sg}, rather than open-ended legal reasoning and interaction with users.

To address these gaps, we introduce \benchname, a benchmark that contains 1{,}100 lawyer-authored source questions across 10 domains of Singapore law: 340 fully specified (FS) and 760 underspecified (US). For each US question, the contributor specifies a fixed assignment of values to the omitted facts and writes a paired FS counterpart that states those facts upfront, yielding 1{,}860 evaluation instances: 1{,}100 FS and 760 US. Each question carries a lawyer-authored rubric organized by Issue, Rule Recall, Application, and Conclusion (IRAC),\footnote{IRAC is a widely taught organizational framework for legal analysis~\citep{metzler2003importance}. We use its components as a rubric schema; models are not required to use IRAC headings.} and each US question shares its rubric with its FS counterpart. These pairs allow us to compare legal analysis when the same material facts must be elicited versus provided upfront. We evaluate 19 LLMs in closed-book and web-enabled settings. Under a prompt that does not reveal specification status, the model must decide whether to answer or clarify. Then, a tool-gated user simulator adapted from $\tau^2$-Bench~\citep{barres2025tau2bench} reveals a hidden fact only when asked. This design lets us assess whether models recognize when clarification is needed, ask about relevant hidden facts, and produce correct legal analysis within each IRAC component.

Our contributions can be summarized as follows: (1) We introduce \benchname, a large-scale, lawyer-authored benchmark for open-ended legal reasoning and selective clarification across 10 domains of Singapore law. (2) We develop a multi-turn evaluation framework that measures selective clarification and fact elicitation and compares legal analysis across paired US/FS instances. (3) Our evaluation of 19 LLMs reveals substantial headroom even with web search and difficulty recognizing when clarification is necessary.

\section{Dataset}
\label{sec:dataset}

\begin{figure}[t]
\centering
\includegraphics[width=\linewidth]{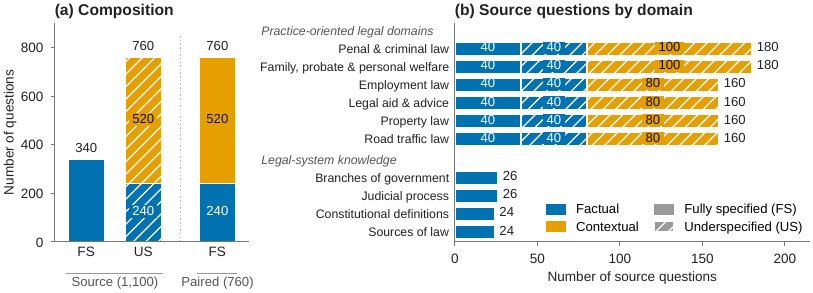}
\caption{Composition of \benchname. (a)~The 1{,}100 source questions comprise 340 FS and 760 US questions. Each US question has a manually written paired FS counterpart, yielding 1{,}860 evaluation instances. (b)~Source questions per domain, excluding the 760 paired counterparts.}
\label{fig:dataset}
\end{figure}

\subsection{Dataset Description}
\label{sec:description}

\benchname contains 1{,}100 English-language source questions about Singapore law, each requiring a free-form response and accompanied by a reference answer and grading rubric (\S\ref{sec:rubrics}). The questions span 10 domains, grouped into six practice-oriented legal domains and four legal-system knowledge domains (Figure~\ref{fig:dataset}b).

Each source question is labeled by \textbf{category} and \textbf{specification}. Factual questions (580) ask for a legal rule, definition, process, or institutional fact, whereas Contextual questions (520) require applying the law to a described situation. FS questions (340) include all contributor-identified facts needed to determine the answer, whereas US questions (760) omit at least one legally material variable. By construction, all 520 Contextual source questions are US; the 580 Factual questions comprise 340 FS and 240 US. Each US source question has a manually written paired FS counterpart, yielding 1{,}860 evaluation instances: 760 matched US--FS pairs and 340 additional FS questions (Figure~\ref{fig:dataset}a). Each pair shares the same underlying legal content as well as rubric, enabling paired comparisons under identical grading criteria with facts available upfront versus through interaction. 

\subsection{Dataset Construction}
\label{sec:collection}

\textbf{Contributors. }
Questions were authored by 23 contributors, all verified lawyers (2 to 37 years of experience, median 6) who held either a Singapore law degree, or an overseas law degree together with active legal practice in Singapore. All completed project-specific quality training. For each item, contributors were assigned a category, domain, subtopic, and specification label, and could decline assignments outside their expertise. Every item was independently reviewed by one of seven contributors. Reviewer edits were final. More details about the contributors are in Appendix~\ref{app:contributors}.

\textbf{Authoring workflow. }
Contributors were instructed to author questions to resemble requests that a layperson might pose to an LLM assistant. For each question, the contributor supplied the question text, reference answer, supporting legal references, and grading rubric (\S\ref{sec:rubrics}). For each US question, the contributor identified the omitted legally material variables and provided, for each variable, a clarifying question and corresponding answer (CQA). We call the joint assignment of these hidden fact values the point-instance. The point-instance defines a single completed fact pattern with a determinate answer under the benchmark rubric. Figure~\ref{fig:example} illustrates this construction, including an US question, the CQAs for the omitted variables, the paired FS counterpart, and the shared rubric.

US questions omit between 1 and 16 variables (median 3; mean 3.1). Fixing one instance keeps the reference answer and rubric focused on a single fact pattern, while the interactive evaluation still tests which missing variables the model asks about. The contributor then manually created the paired FS counterpart by incorporating the point-instance facts while preserving the underlying question. We will release 50\% of source questions in each category and retain the remaining 550 for private evaluation, keeping each US-FS pair in the same split.

\begin{table}[t]
\centering
\small
\setlength{\tabcolsep}{4pt}
\begin{tabularx}{\linewidth}{@{}lXrrrr@{}}
\toprule
 & & \multicolumn{3}{c}{\% of mandatory items} & \\
\cmidrule(lr){3-5}
Dimension & Credits the model for & Factual & Contextual & All & \% Mand. \\
\midrule
Issue       & Identifying a relevant legal issue or variable & 13.1 & 24.3 & 19.0 & 98.6 \\
Rule Recall & Stating or correctly describing the governing law                & 58.5 & 40.6 & 49.2 & 82.5 \\
Application & Applying the rule to the facts of the instance                  & 14.6 & 21.4 & 18.2 & 94.0 \\
Conclusion  & Reaching the correct legal outcome                              & 13.8 & 13.7 & 13.7 & 96.0 \\
\midrule
\multicolumn{2}{@{}l}{Mandatory items per source question (mean)}              &  9.2 & 11.2 & 10.1 &      \\
\bottomrule
\end{tabularx}
\caption{IRAC rubric dimensions and their distribution over mandatory items (Factual n{=}5{,}329; Contextual n{=}5{,}833; all n{=}11{,}162). \% Mand. is the share of each dimension tagged Mandatory.}
\label{tab:dimensions}
\end{table}

\subsection{IRAC Rubrics}
\label{sec:rubrics}

Each evaluation instance is graded against a rubric of 1--23 binary (satisfied/not satisfied) items. Across the 1{,}100 source questions, contributors authored 12{,}546 distinct rubric items. Applying each shared rubric to both members of its US-FS pair yields 21{,}762 instance-level rubric items in total (mean 11.7 per evaluation instance). Contributors were instructed to make each rubric item atomic, self-contained, and non-overlapping. Each item carries two tags: \textbf{dimension} and \textbf{importance}.

\textbf{Dimensions. } 
Following recent rubric-based legal evaluations~\citep{fan2026lexam,chlapanis2025greekbarbench,lee2026evaluating,shi2026plawbench}, we score responses using itemized criteria. We organize these criteria into four IRAC-inspired dimensions~\citep{metzler2003importance}: Issue, Rule Recall, Application, and Conclusion. This scheme evaluates distinct parts of the model's expressed legal analysis, and can localize errors such as correct rule recall followed by flawed application. Not every rubric contains items from all four dimensions. Table~\ref{tab:dimensions} defines the dimensions and reports their distribution.

\textbf{Importance.  }
Of the 12{,}546 distinct rubric items, 11{,}162 (89.0\%) are Mandatory, meaning they are necessary for a substantively correct response. The remaining 1{,}384 (11.0\%) are Optional and capture useful but nonessential content, such as citing a specific statute when no citation is requested or discussing additional requirements after one condition has already determined the outcome. Headline results in \S\ref{sec:results} are computed over mandatory items only.

\section{Experiments}
\label{sec:experiments}

\subsection{Models}
\label{sec:models}

We evaluate the same 19 general-purpose LLMs in closed-book and web-search conditions. Open-weight models are grouped by total parameter count. \textbf{Proprietary models} (9): Claude Fable~5~\citep{anthropic2026fable}, Claude Opus~4.8~\citep{anthropic2026opus48}, and Claude Sonnet~5~\citep{anthropic2026sonnet5}; GPT-5.6 Sol, Terra, and Luna~\citep{openai2026gpt56}; Gemini~3.1~Pro~\citep{googledeepmind2026gemini31pro} and Gemini~3.5~Flash~\citep{googledeepmind2026gemini35flash}; and Qwen3.7~Plus~\citep{qwen2026qwen37plus}.
\textbf{Larger open-weight models} (6): DeepSeek-V4~Pro and Flash~\citep{deepseek2026v4}; Kimi~K2.6~\citep{moonshot2026kimik26} and Kimi~K3~\citep{moonshot2026kimik3}; Mistral~Large~3~\citep{mistral2025mistral3}; and Llama~4~Scout~\citep{meta2025llama4}.
\textbf{Smaller open-weight models} (4): Qwen3.6~27B~\citep{qwen2026qwen36} and Qwen-SEA-LION-v4.5-27B-IT~\citep{aisingapore2026sealion45}; Gemma~3~27B~IT~\citep{gemmateam2025gemma3} and Gemma-SEA-LION-v4-27B-IT~\citep{aisingapore2025sealion4}.
The two SEA-LION models are Southeast Asia-adapted members of AI~Singapore's SEA-LION family~\citep{ng2025sealion}.
Their respective base models, Qwen3.6~27B and Gemma~3~27B~IT, are included for comparisons of regional adaptation. Access routes, reasoning and sampling configurations, and output-token ceilings are detailed in Appendix~\ref{app:models}.

\subsection{Multi-Turn Evaluation Pipeline}
\label{sec:pipeline}

\begin{figure}[t]
    \centering
    \includegraphics[width=\linewidth]{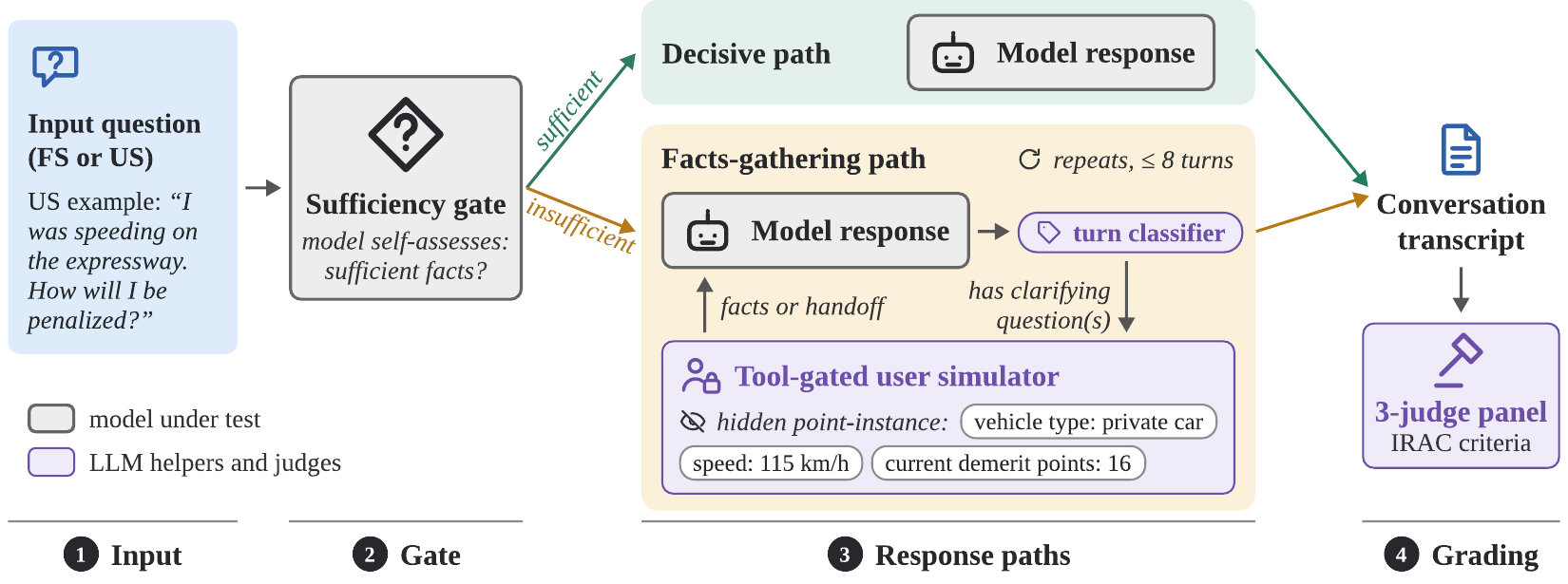}
    \caption{Two-stage sufficiency-gated evaluation. The model under test selects a decisive or facts-gathering path at the gate. A panel of three LLM judges grades the transcript against IRAC rubrics.}
    \label{fig:pipeline}
\end{figure}

Figure~\ref{fig:pipeline} illustrates our experiment pipeline. Every FS and US input question uses the same two-stage sufficiency-gated prompt scheme (Appendix~\ref{app:prompts}) without disclosing specification status. We run closed-book and web-search conditions, with search available only during the evaluated model's dialogue turns. The gate and helpers remain closed-book.

\textbf{Sufficiency gate. }
Given a question, a separate call asks the evaluated model to output \texttt{SUFFICIENT\_FACTS} or \texttt{INSUFFICIENT\_FACTS}. The harness then supplies only the chosen path's system prompt. This decoupling makes the sufficiency judgment an explicit, directly measurable decision, separate from the model’s subsequent behavior of either clarifying or answering. We keep this gate closed-book so that routing reflects the model's judgment from the question itself and remains comparable across retrieval conditions, rather than depending on retrieved evidence. Appendix~\ref{app:metrics} reports how often models actually request facts after being routed to facts-gathering.

\textbf{Response paths. }
If the model selects sufficient facts, it enters the Decisive path and is instructed to provide direct legal analysis and a decisive conclusion without clarifying questions. The conversation ends after its first response, without invoking the user simulator. Otherwise, the model enters the Facts-gathering path and interacts with the user simulator for up to eight model turns. An LLM turn classifier (Opus 4.6) detects whether there are clarifying questions about concrete facts in the model response. If not, the classifier ends the conversation. Conversations always end on a model response. The model is instructed to output \texttt{INSUFFICIENT\_FACTS\_GATHERED} rather than guess when facts remain insufficient.

\textbf{Tool-gated user simulator. }
Following tool-constrained user simulation~\citep{barres2025tau2bench,trinh2026hil}, we build a simulator (Opus 4.8) that maps fact requests in the latest model response to ground-truth CQAs, selects the variables that are explicitly asked, and rephrases the answers into natural sentences without adding new information. When two consecutive model responses match no fact variable, a decisive handoff is triggered that permits one final model turn within the turn limit. FS questions trigger this handoff immediately. Simulator-selected variables are recorded for each question. 

Validated in \S\ref{sec:human-validation}, the LLM turn classifier and user simulator run without reasoning under fixed configurations (Appendix~\ref{app:models}). All reported rubric scores use criterion-level majority votes from the selected three-model panel (\S\ref{sec:judge}). The planned grid is $3\times19\times1{,}860=106{,}020$ generation trials per condition, or 212{,}040 total, excluding validation. 

\subsection{Metrics}
\label{sec:metrics}

Our primary metric, \textbf{mandatory score}, is the percentage of mandatory rubric criteria passed in a trial. It is macro-averaged across trials for each model and condition, overall and separately for FS and US questions. Optional rubric criteria are excluded. 

Multiple metrics characterize different aspects of a model's clarification behavior. \textbf{Gate selectivity} compares the recorded route at the sufficiency gate with the specification label. \textbf{Clarification rate} records at least one concrete fact request. 
On US trials, the user simulator records which hidden variables were requested by the model. For each model and run, let $m_i$ be the number of variables requested, $v_i$ the total hidden variables, and $N_{\mathrm{US}}$ the number of US trials. Task-averaged \textbf{variable coverage} is $\frac{100}{N_{\mathrm{US}}}\sum_{i=1}^{N_{\mathrm{US}}} m_i/v_i$, including trials with no requests ($m_i=0$), weighting trials equally rather than pooling variables. 

\subsection{LLM-as-Judge Meta-Evaluation}
\label{sec:judge}

\textbf{Human reference. }
We stratify 110 source tasks across the taxonomy: 34 FS-only tasks and 76 US--FS pairs, yielding 186 evaluation instances. Opus~4.8, DeepSeek-V4~Flash, and Qwen-SEA-LION-v4.5-27B-IT generate 558 closed-book conversations with reasoning enabled. Two reviewers grade each response against binary IRAC criteria, without adjudication or label merging. Across all 110 tasks (6{,}522 paired criterion items), raw agreement is 80.4\% and Fleiss' $\kappa=0.607$ (95\% CI $[0.569,0.643]$; question-cluster bootstrap).

\textbf{Judge-panel selection. }
Nineteen judges use a shared frozen grading prompt derived from the human guidelines, with reasoning and search disabled. We enumerate all 969 three-judge panels, take criterion-level majority votes, and rank panels by mean Cohen's $\kappa$ against the two annotation copies. The top judge panel---\textbf{DeepSeek-V4~Flash, GPT-5.6~Sol, and Kimi~K3}---scores $\kappa=0.697$ and 84.8\% agreement on this reference. These are in-sample estimates. This selected panel grades both production conditions.

Grouped five-fold cross-validation on the same reference yields pooled held-out $\kappa=0.690$ and 84.5\% agreement, compared with $\kappa=0.678$ for single-judge selection. Panels and individual judges are selected within training folds, keeping each source's variants, responses, and criteria together. This evaluates panel selection, not the fixed production panel. Appendix~\ref{app:judge-meta} details the reference and protocol.

\subsection{Additional Human Validation of LLM Helpers}
\label{sec:human-validation}

We audit fact-request recognition by the LLM turn classifier and faithful disclosure of requested facts by the LLM user simulator. Two independent annotations, from three contributors, cover each of 55 tasks: 17 FS-only tasks and 38 US--FS pairs. Annotators agree perfectly on which facts were disclosed (Fleiss' $\kappa=1.00$) and substantially on which were requested ($\kappa=0.77$); agreement on whether a turn warrants a simulator reply is $\kappa=0.82$ for 64 FS turns and $0.71$ for 38 US conversations. Across 64 bank facts in 24 US conversations with simulator replies, the simulator achieves \textbf{89.1\% fact-level disclosure accuracy} (93.8\% and 84.4\% by annotation copy), where a requested fact should be disclosed and an unrequested fact withheld. Among these 24 conversations, both annotators identify over-disclosure in four (16.7\%), with five additional disputed cases; neither identifies under-disclosure. Both reviewers label as fact requests all nine FS turns whose positive classifier decisions can be reconstructed and matched to turn-specific human labels. Appendix~\ref{app:human-validation} details the scoring, alignment, and annotation reliability.

\section{Results}
\label{sec:results}

\begin{figure}[!htbp]
\setlength{\abovecaptionskip}{4pt}
\centering
\begin{minipage}[t]{0.49\textwidth}
\vspace{0pt}
\centering
\begin{minipage}[c][2.85in][c]{\linewidth}
\centering
\begingroup
\fontsize{8}{9.2}\selectfont
\setlength{\tabcolsep}{2.5pt}
\renewcommand{\arraystretch}{1.0}
\ifdefined\mainresultstablebox\else\newsavebox{\mainresultstablebox}\fi
\sbox{\mainresultstablebox}{%
\begin{tabular}{@{}clrr@{}}
\toprule
& Model & Closed book & Web enabled\\
\midrule
\multirow{9}{*}{\scalebox{0.8}{\rotatebox[origin=c]{90}{Proprietary}}} & Fable 5 & $\mathbf{69.8}\!\pm\!0.5$ & $73.3\!\pm\!0.1$ \\
 & Opus 4.8 & $66.6\!\pm\!0.2$ & $72.9\!\pm\!0.2$ \\
 & GPT-5.6 Sol & $64.6\!\pm\!0.3$ & $70.5\!\pm\!0.3$ \\
 & Gemini 3.5 Flash & $62.5\!\pm\!0.4$ & $71.6\!\pm\!0.4$ \\
 & Sonnet 5 & $61.2\!\pm\!0.2$ & $70.1\!\pm\!0.3$ \\
 & GPT-5.6 Terra & $59.8\!\pm\!0.4$ & $67.1\!\pm\!0.1$ \\
 & GPT-5.6 Luna & $58.4\!\pm\!0.3$ & $66.7\!\pm\!0.1$ \\
 & Gemini 3.1 Pro & $57.2\!\pm\!0.1$ & $59.9\!\pm\!0.7$ \\
 & Qwen3.7 Plus & $49.9\!\pm\!0.5$ & $67.9^{*}\!\pm\!0.6$ \\
\midrule
\multirow{6}{*}{\scalebox{0.8}{\rotatebox[origin=c]{90}{\shortstack{Larger\\open-weight}}}} & Kimi K3 & $65.9\!\pm\!0.3$ & $\mathbf{77.8}\!\pm\!1.2$ \\
 & Kimi K2.6 & $54.2\!\pm\!0.4$ & $68.9\!\pm\!0.7$ \\
 & DeepSeek V4 Pro & $52.4\!\pm\!0.1$ & $73.1\!\pm\!1.1$ \\
 & Mistral 3 Large & $51.8\!\pm\!0.2$ & $57.2\!\pm\!0.2$ \\
 & DeepSeek V4 Flash & $50.1\!\pm\!0.3$ & $74.0\!\pm\!0.4$ \\
 & Llama 4 Scout & $36.0\!\pm\!0.4$ & $37.5^{\dagger}$ \\
\midrule
\multirow{4}{*}{\scalebox{0.8}{\rotatebox[origin=c]{90}{\shortstack{Smaller\\open-weight}}}} & Qwen3.6 27B & $47.0\!\pm\!0.4$ & $58.9\!\pm\!0.3$ \\
 & Qwen-SEA-LION v4.5 27B & $46.1\!\pm\!0.3$ & $54.7\!\pm\!1.1$ \\
 & Gemma 3 27B & $35.0\!\pm\!0.3$ & $37.7\!\pm\!0.4$ \\
 & Gemma-SEA-LION v4 27B & $34.5\!\pm\!0.1$ & $37.3\!\pm\!0.4$ \\
\bottomrule
\end{tabular}%
}
\ifdim\wd\mainresultstablebox>\linewidth
  \sbox{\mainresultstablebox}{\resizebox{\linewidth}{!}{\usebox{\mainresultstablebox}}}%
\fi
\ifdim\dimexpr\ht\mainresultstablebox+\dp\mainresultstablebox\relax>2.85in
  \resizebox*{!}{2.85in}{\usebox{\mainresultstablebox}}%
\else
  \usebox{\mainresultstablebox}
\fi
\endgroup
\end{minipage}
\makeatletter
\ifdefined\captionof
  \def\resultstablecaption{\captionof{table}}
\else
  \def\@captype{table}
  \let\resultstablecaption\caption
\fi
\makeatother
\resultstablecaption{\textbf{Mandatory scores (\%), FS+US combined.} Model rows show mean $\pm$ sample SD across three seeds; SD is estimated from rounded seed scores. Rows are sorted by Closed book score within each group.}
\label{tab:main-results}
\vspace{2pt}
\end{minipage}\hfill
\begin{minipage}[t]{0.49\textwidth}
\vspace{4pt}
\begin{minipage}[c][2.85in][c]{\linewidth}
\includegraphics[width=\linewidth]{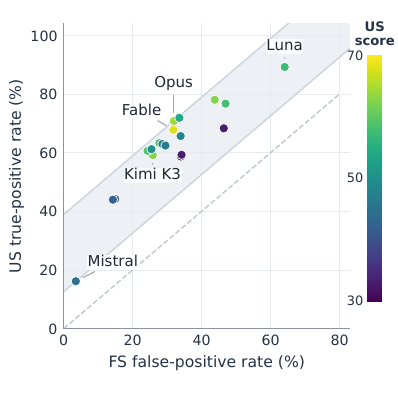}
\end{minipage}
\caption{\textbf{Closed-book gate selectivity.} Each point represents one model. Color denotes mandatory score across all US trials. TPR and FPR are the proportions of US and FS trials routed to fact gathering. Shading marks the observed TPR-FPR band.}
\label{fig:results-gate}
\end{minipage}
\end{figure}

\subsection{Web search improves performance but leaves substantial headroom}
\label{sec:results-overall}

Table~\ref{tab:main-results} reports mandatory scores for all 19 models, combining FS and US questions. Fable~5 leads closed book at \textbf{69.8\%}; Kimi~K3 leads the web-enabled condition at \textbf{77.8\%}. All 19 models improve with web search, but gains vary substantially: DeepSeek~V4 Flash gains the most, rising from 50.1\% to 74.0\% (+23.9 percentage points). The mean across models rises from 53.8\% to 63.0\% (+9.2 percentage points), \textbf{leaving substantial headroom even with web search enabled}. Rule Recall remains the weakest IRAC component (43.2\% closed book; 56.5\% web-enabled). Appendix~\ref{app:additional-results} disaggregates performance by task type, specification status, IRAC component, and legal domain, and reports model-level search use and per-model clarification behavior.

\subsection{Sufficiency recognition limits selective clarification}
\label{sec:results-selectivity}

The closed-book gate routes 37.8\% of US trials to the decisive path and 31.3\% of FS trials to fact gathering. Figure~\ref{fig:results-gate} shows that models cluster within a narrow diagonal band, $12.6\leq\mathrm{TPR}-\mathrm{FPR}\leq38.9$ percentage points. Detection and false-positive rates tend to rise together across models ($r=0.90$): Luna detects 89.3\% of US trials but flags 64.1\% of FS trials, whereas Mistral detects only 16.2\% with 3.5\% false positives. Even the best balanced gate accuracy, weighting FS and US equally, is only 69.4\% (Opus). \textbf{Greater sensitivity to missing facts thus accompanies more unnecessary routing to fact gathering}: models struggle to distinguish queries that need clarification from those already answerable. Evaluating clarification as an active decision exposes a failure mode that evaluations presupposing a need to clarify would miss.

\subsection{Targeted clarification is associated with smaller missing-fact penalties}
\label{sec:results-paired}

\begin{figure}[!htbp]
\setlength{\abovecaptionskip}{4pt}
\centering
\includegraphics[width=\textwidth]{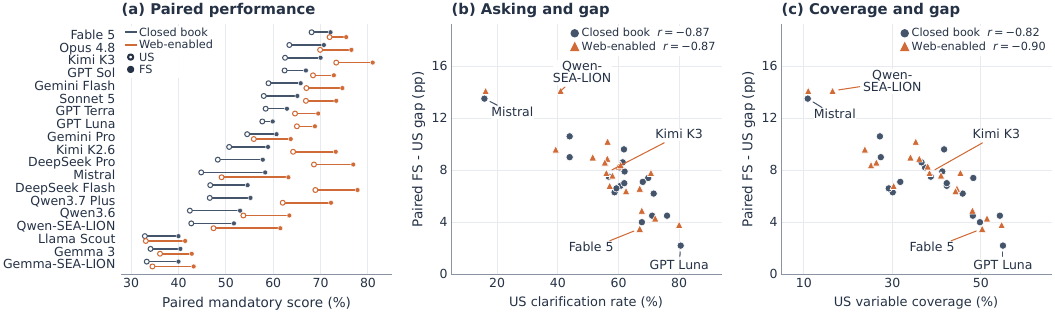}
\caption{\textbf{Paired performance and clarification.} (a) Mean mandatory scores on 760 matched US-FS pairs. Each segment joins US and FS for one model and condition. (b-c) Paired FS--US gap versus US clarification rate and target-request coverage, respectively. Each point represents one model and condition. Pearson $r$ is computed across the 19 models within each condition.}
\label{fig:results-paired}
\end{figure}

Figure~\ref{fig:results-paired}(a) compares 760 matched US-FS pairs, holding the source questions constant. Every model scores higher on FS on average in both conditions. With web search, the gap appears to widen for 13 of 19 models, though we treat this as a descriptive trend. Thus, higher web-enabled scores do not imply that models have overcome underspecification. Models that clarify more frequently tend to have smaller paired gaps ($r=-0.87$ in both conditions; Figure~\ref{fig:results-paired}(b)), as do those requesting a larger share of target facts ($r=-0.82$ closed book; $r=-0.90$ web-enabled; Figure~\ref{fig:results-paired}(c)).

For absolute US performance, \textbf{what models ask is more informative than asking frequency alone} (Figure~\ref{fig:results-outcomes}(c-d)): in closed book, target-request coverage correlates more strongly with US score than clarification rate does ($r=0.73$ versus $0.40$), and the web-enabled condition sees a similar pattern ($r=0.67$ versus $0.35$). These model-level associations do not establish causal effects: coverage measures facts requested, not verified receipt or successful use, and complete coverage need not be necessary when one fact determines the outcome.

\begingroup
\renewcommand{\thefootnote}{}
\footnotetext[0]{%
$^{*}$Two-seed mean.
$^{\dagger}$One seed; no SD.
Both models were deprovisioned mid-run, so only partial
data are available and reported.}
\endgroup

\subsection{Conclusion-rubric and non-conclusions reveal distinct failure modes}
\label{sec:results-conclusions}

\begin{figure}[!htbp]
\setlength{\abovecaptionskip}{4pt}
\centering
\includegraphics[width=\textwidth]{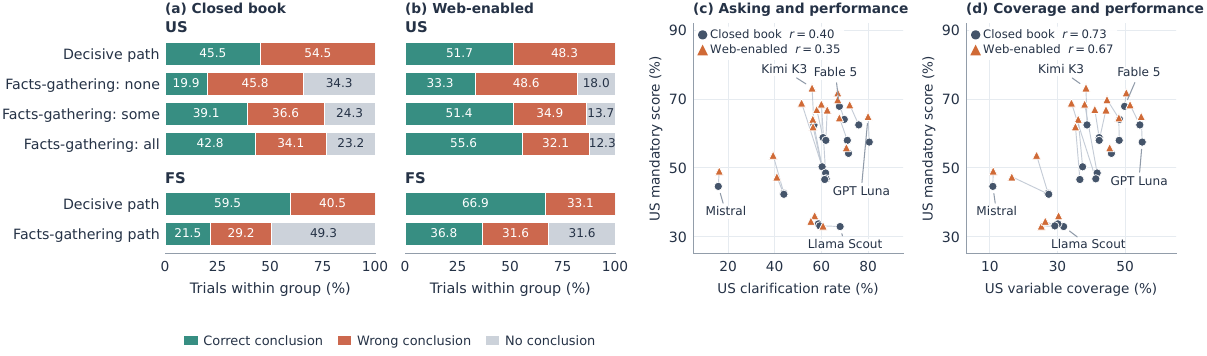}
\caption{\textbf{Clarification and performance.} (a-b) Recorded conclusion outcomes by route and target-request coverage. Percentages normalize within rows. ``Correct conclusion'' and ``Wrong conclusion'' denote Conclusion-rubric pass/fail among responses treated as committed. ``Decisive path'' and ``Facts-gathering'' identify recorded routes; for US facts-gathering rows, ``none/some/all'' refers to target variables detected as requested. (c-d) US mandatory score versus US clarification rate and target-request coverage, respectively. Color and shape denote condition; segments connect the same model across conditions. Pearson $r$ is computed across the 19 models within each condition.}
\label{fig:results-outcomes}
\end{figure}

Figure~\ref{fig:results-outcomes}(a--b) separates rubric-passing and rubric-failing responses treated as committed from detected non-conclusions. Among closed-book US facts-gathering trials, the rubric-failing share decreases with greater target-request coverage: 45.8\%, 36.6\%, and 34.1\% for zero, partial, and complete coverage. Complete-coverage trials also have a smaller rubric-failing share than direct-answer trials (34.1\% versus 54.5\%), though 23.2\% are detected as non-conclusions, leaving rubric-passing shares similar (42.8\% versus 45.5\%). Among FS trials routed to fact gathering, 49.3\% are detected as non-conclusions, despite the required facts already being supplied.

With web search, detected non-conclusion rates are lower across all facts-gathering groups, but rubric-failing shares do not consistently decline. For complete-coverage US trials, the rubric-passing share rises to 55.6\%: detected non-conclusions fall from 23.2\% to 12.3\%, whereas the rubric-failing share declines only from 34.1\% to 32.1\%. This failing share instead rises slightly in the zero-target US group (45.8\% to 48.6\%) and the FS facts-gathering group (29.2\% to 31.6\%). The more consistent change is therefore in detected non-conclusions, with rubric-failing responses remaining common. Group membership varies across routes and conditions, so these comparisons are descriptive.

\section{Related Work}
\label{sec:related}

\subsection{Legal Benchmarks for LLMs}

A growing family of benchmarks evaluates LLMs on jurisdiction-specific law. LegalBench~\citep{guha2023legalbench} collaboratively assembles tasks spanning six types of legal reasoning, primarily over U.S. law. For Chinese law, LawBench~\citep{fei2024lawbench} tests legal tasks across three cognitive levels, LexEval~\citep{li2024lexeval} covers a broader range of legal tasks, LAiW~\citep{dai2025laiw} organizes its tasks around legal syllogism, and LegalAgentBench~\citep{li2025legalagentbench} evaluates tool-using agents. Exam-derived benchmarks include LEXam~\citep{fan2026lexam}, built from Swiss, European, and international law exams, and GreekBarBench~\citep{chlapanis2025greekbarbench}, which evaluates free-text answers grounded in statutes and case facts from Greek Bar examinations. Further work covers Korea~\citep{kim2024koreanlegal,lee2025koblex}, Thailand~\citep{akarajaradwong2025nitibench}, and India~\citep{joshi2024iltur}, as well as Arabic-language legal evaluation~\citep{hijazi2024arablegaleval}. For Singapore law, AutoLaw studies compliance and violation detection~\citep{nguyen2025autolaw}, MLEB evaluates legal embeddings for retrieval~\citep{butler2025mleb}, and SG-LegalCite evaluates citation retrieval from Singapore judgments~\citep{ern2026sg}. \benchname complements these benchmarks by evaluating open-ended legal analysis and selective clarification for Singapore law. A parallel line of work develops legal-domain-adapted LLMs, including SaulLM~\citep{colombo2024saullm}, ChatLaw~\citep{cui2023chatlaw}, and DISC-LawLLM~\citep{yue2023disclawllm}. Our experiments instead focus on contemporary general-purpose models and regional models for Southeast Asia, evaluating their ability to handle Singapore-specific reasoning without legal-domain adaptation.

Recent legal benchmarks begin to evaluate multi-turn consultation and fact elicitation. LexRAG~\citep{li2025lexrag} evaluates answers over pre-authored consultation dialogues, while PLawBench's public legal consultation task~\citep{shi2026plawbench} mandates a one-shot list of follow-up questions without simulating responses or grading subsequent legal analysis. LeCoDe~\citep{yuan2026lecode} lets a lawyer agent choose between questions and advice, evaluating key-fact recall and advice quality. FSDBench~\citep{bystronski2026factual} evaluates multi-turn fact elicitation in Polish tax law without assessing the resulting legal answer. Closest to our setting, DLawBench~\citep{zhang2026dlawbench} lets a model choose between further client questions and submitting a legal memorandum, assessing inquiry and legal analysis. \benchname complements this literature by testing whether to clarify across mixed fully specified and underspecified inputs. Paired questions share a fixed fact assignment and rubric, allowing comparison of legal analysis when material facts must be elicited versus provided upfront.

\subsection{Underspecification and Clarification in LLM Evaluation}

Outside the legal domain, CLAMBER~\citep{zhang2024clamber} finds that LLMs struggle to detect and clarify ambiguity, while AbstentionBench~\citep{kirichenko2025abstentionbench} reports abstention failures on unanswerable and underspecified questions. QuestBench studies which clarification questions resolve underspecified reasoning tasks~\citep{li2026questbench}, while Clarify When Necessary~\citep{zhang2025clarify} and AskBench~\citep{zhao2026ask} study when and what models should ask. ClarifyMT-Bench~\citep{luo2025clarifymt} and RegretBench~\citep{ta2026one} evaluate multi-turn clarification policies. LHAW~\citep{pu2026lhaw} and HiL-Bench~\citep{trinh2026hil} address underspecified reasoning in long-horizon agentic tasks. For user simulation, we adapt $\tau^2$-Bench's use of tools and observable state to constrain our LLM-based user simulator~\citep{barres2025tau2bench}.
In legal consultation, underspecification is conditioned by the governing doctrine. 
\benchname evaluates this connection between jurisdiction-specific legal knowledge and selective clarification.

\section{Limitations}
\label{sec:limitations}
Our study is subject to a set of limitations. First, human validation of the LLM helpers is limited in scope, and turn-classification errors or simulator over-disclosure may affect dialogue trajectories and downstream scores. Second, provider differences prevent identical search-tool access, so web-enabled results reflect both model capabilities and tool configurations. While the decoupled, closed-book gate makes sufficiency judgments explicit and supports controlled comparisons across retrieval conditions, its up-front routing may alter clarification behavior, reducing ecological validity. Finally, not every annotated variable is necessary to resolve a point-instance: a subset of facts may be dispositive, so incomplete target-request coverage need not indicate inadequate clarification. 

\section{Conclusion}
\label{sec:conclusion}

We introduced SG-LegalClarify, a benchmark for selective clarification in Singapore legal reasoning that combines fully specified and underspecified questions, a fact-grounded user simulator, and rubric-based evaluation. Our evaluation of 19 models reveals persistent weaknesses in recognizing when clarification is needed and conducting correct legal analysis. Web search improves overall scores, but fact-gathering groups show more consistent reductions in non-conclusions than in rubric-failing conclusions. Together, these findings motivate evaluating legal assistants beyond answer quality or clarification frequency alone. SG-LegalClarify provides a testbed for jointly improving when models ask, what they ask, and how correctly they answer.

\section*{AI use statement}
In this work, we used generative AI tools for coding and editing paper drafts more efficiently. We have reviewed all AI-assisted work.  We take responsibility for the final content of this work, including text, claims or artifacts produced with the aid of generative AI.

\section*{Ethics Statement}

\benchname is intended to support research on reliable legal reasoning and clarification. Questions and rubrics were authored and reviewed by verified lawyers, who could decline assignments outside their expertise. Sensitive-topic assignments were restricted to contributors who opted in. Incorrect legal guidance can have serious consequences, and benchmark scores should not be interpreted as evidence that a model is safe for unsupervised legal advice.

\section*{Reproducibility Statement}

We describe dataset construction, rubric design, contributor workflow, and review procedures. We further document the evaluation protocol, model and tool configurations, seed handling, metrics, and all system prompts for easier reproduction of our work. We will release the public benchmark split, full evaluation pipeline code, and model outputs on the public split. 

\bibliographystyle{abbrvnat}
\bibliography{references}

\clearpage
\appendix

\section{Contributor Details}
\label{app:contributors}

\paragraph{Recruitment and qualifications.}
All contributors were verified lawyers with demonstrated expertise in Singapore law. Two groups took part (Table~\ref{tab:contributors}): 23 contributors authored and reviewed benchmark items, and nine provided the human labels used to validate the LLM judges (Section~\ref{sec:judge}). Items involving sensitive subject matter were assigned only to contributors who had opted in to such content.

\begin{table}[h]
\centering
\small
\begin{tabular}{lrrrrr}
\toprule
Pool & $N$ & Mean & Median & Min & Max \\
\midrule
Question authors and reviewers & 23 & 9.2 & 6 & 2 & 37 \\
Meta-evaluation annotators & 9 & 5.6 & 4 & 2 & 13 \\
\bottomrule
\end{tabular}
\caption{Contributor pools and years of legal experience.}
\label{tab:contributors}
\end{table}

\paragraph{Review.}
Seven of the 23 authors also served as reviewers. The review covered (i) the specification label (fully specified or underspecified), (ii) the accuracy of the clarifying question and answer pairs, (iii) the legal accuracy of the expected answer, (iv) the accuracy of the listed legal references, (v) the rubrics, and (vi) the alignment of the category and domain labels with the item. Reviewers could edit any field or relabel the item. The reviewed version superseded the author's and was final, with no further adjudication.

\paragraph{Meta-evaluation annotators.}
Nine lawyers (2 to 13 years of experience, median 4) produced the human labels for candidate-judge evaluation (Section~\ref{sec:judge}). Each response and rubric item received two independent binary judgments. Panel selection compared predictions separately with both annotation copies and averaged the two Cohen's $\kappa$ values; no majority human reference or adjudication was used. Annotator identities can vary across questions.

\section{Model and Inference Configuration}
\label{app:models}

Tables~\ref{tab:model-config} and~\ref{tab:web-search-config} list the 19 models' closed-book configurations and web-search routes and controls, respectively. For closed-book calls, supported models are requested to reason at high effort; Qwen-SEA-LION uses template-based thinking. The closed-book output ceiling is 16{,}384 tokens per call, except for Llama~4~Scout (8{,}192). Available reasoning outputs are logged alongside conversations and run metadata.

\paragraph{Routing.}
\begingroup\sloppy
Gemini~3.1~Pro uses \path{gemini/gemini-3.1-pro-preview}; Mistral~3~Large uses \path{mistral/mistral-large-latest}; and Qwen3.6~27B uses \path{fireworks_ai/deployed/t5cvm6h4}. Llama~4~Scout uses \path{bedrock/us.meta.llama4-scout-17b-instruct-v1:0} closed-book and \path{groq/meta-llama/llama-4-scout-17b-16e-instruct} for its retained web-search run. Gemma~3~27B~IT uses \path{bedrock/google.gemma-3-27b-it} closed-book, but switches to local \path{gemma-3-27b-it} for web search because the Bedrock route rejects the search-tool schema.
\par\endgroup

\paragraph{Web-search configuration.}
The eight OpenAI, Anthropic, and Gemini models use native provider search through the OpenAI Responses API, Anthropic citations, and Gemini grounding, respectively; the remaining 11 use a custom Serper--Jina search-and-visit loop. Native search uses a medium context setting, with Anthropic limited to 100 search-tool uses per API call; OpenAI and Gemini use provider-managed limits. The custom loop permits up to 999 tool-call rounds, 9{,}999 searches, and 9{,}999 page visits per assistant turn. The models otherwise retain closed-book sampling settings shown in Table~\ref{tab:model-config}.

\begin{table}[t]
\centering
\small
\setlength{\tabcolsep}{4pt}
\begin{tabularx}{\linewidth}{@{}Xllr@{}}
\toprule
Model & Closed-book route & Reasoning & Temp. \\
\midrule
Claude Fable 5 & Anthropic & adaptive, high & 1.0 \\
Claude Opus 4.8 & Anthropic & adaptive, high & 1.0 \\
Claude Sonnet 5 & Anthropic & adaptive, high & 1.0 \\
GPT-5.6 Sol & OpenAI & high & 1.0 \\
GPT-5.6 Terra & OpenAI & high & 1.0 \\
GPT-5.6 Luna & OpenAI & high & 1.0 \\
Gemini 3.1 Pro & Google & high & default \\
Gemini 3.5 Flash & Google & high & default \\
Qwen3.7 Plus & Fireworks & high & 0.7 \\
Kimi K2.6 & Fireworks & high & 0.6 \\
Kimi K3 & Fireworks & high & 0.6 \\
DeepSeek-V4 Pro & Fireworks & high & 0.6 \\
DeepSeek-V4 Flash & Fireworks & high & 0.6 \\
Mistral 3 Large & Mistral & --- & 0.0 \\
Qwen3.6 27B & Fireworks (private) & high & 0.7 \\
Llama 4 Scout & AWS Bedrock & --- & 0.0 \\
Qwen-SEA-LION-v4.5-27B-IT & local vLLM & Qwen3 thinking & 1.0 \\
Gemma-SEA-LION-v4-27B-IT & local vLLM & --- & 1.0 \\
Gemma 3 27B IT & AWS Bedrock & --- & 0.0 \\
\bottomrule
\end{tabularx}
\caption{Closed-book inference settings. High effort is requested through the provider-specific reasoning interface; --- denotes no reasoning mode. Gemini temperature is unset. The client retries without temperature when a provider rejects it. SEA-LION sampling uses top-$p$ 0.95, with top-$k$ 20 for Qwen and 64 for Gemma; Qwen additionally uses min-$p$ 0 and repetition penalty 1.}
\label{tab:model-config}
\end{table}

\begin{table}[t]
\centering
\small
\setlength{\tabcolsep}{4pt}
\begin{tabularx}{\linewidth}{@{}Xlll@{}}
\toprule
Model & Web-search route & Search & Tool request \\
\midrule
Claude Fable 5 & Anthropic & Native & --- \\
Claude Opus 4.8 & Anthropic & Native & --- \\
Claude Sonnet 5 & Anthropic & Native & --- \\
GPT-5.6 Sol & OpenAI & Native & --- \\
GPT-5.6 Terra & OpenAI & Native & --- \\
GPT-5.6 Luna & OpenAI & Native & --- \\
Gemini 3.1 Pro & Google & Native & --- \\
Gemini 3.5 Flash & Google & Native & --- \\
\midrule
Qwen3.7 Plus & Fireworks & Custom & API \\
Kimi K2.6 & Fireworks & Custom & API \\
Kimi K3 & Fireworks & Custom & API \\
DeepSeek-V4 Pro & Fireworks & Custom & API \\
DeepSeek-V4 Flash & Fireworks & Custom & API \\
Mistral 3 Large & Mistral & Custom & API \\
Qwen3.6 27B & Fireworks (private) & Custom & Passthrough \\
Llama 4 Scout & Groq & Custom & Wording \\
Qwen-SEA-LION-v4.5-27B-IT & local vLLM & Custom & Wording \\
Gemma-SEA-LION-v4-27B-IT & local vLLM & Custom & Wording \\
Gemma 3 27B IT & local vLLM & Custom & Wording \\
\bottomrule
\end{tabularx}
\caption{Web-search routes and controls for the retained evaluation runs. Native denotes provider-hosted search; Custom denotes the Serper--Jina loop. Tool requests apply to the first custom-loop round on the facts-gathering path: API forces a tool call via \texttt{tool\_choice}; Passthrough forwards this parameter through LiteLLM; Wording relies on a prompt instruction. --- denotes not applicable. Local vLLM endpoints accept \texttt{tool\_choice}, but the harness uses wording to avoid tool-format leaks in Gemma variants.}
\label{tab:web-search-config}
\end{table}

\paragraph{Seeds and run aggregation.}
The planned design comprises three runs per condition, with $19\times1{,}860=35{,}340$ generation trials per run: 106{,}020 per condition and 212{,}040 overall, including 43{,}320 planned US trials per condition. Metrics are computed within each model/run and then averaged over eligible runs. Following deprovisioning, web-search results retain run~1 for Llama~4~Scout and runs~1--2 for Qwen3.7~Plus. All other models retain three runs per condition. Validation studies and ablations are excluded from these totals.

Run directories 1--3 correspond by repository convention to seeds 42, 43, and 44; individual trial records do not stamp the seed/run ID. The closed-book client forwards seeds only to OpenAI and Mistral routes; the custom web-search client forwards seeds to local vLLM. Helpers are unseeded and use the same configuration across conditions; this does not imply identical generated replies.

\paragraph{Helper configuration.}
The configured defaults are Claude Opus~4.6 for the turn classifier, and Claude Opus~4.8 for simulator selection/rendering. Helpers run without reasoning; the sufficiency gate uses the evaluated model. Reported rubric scores use the selected three-model panel (\S\ref{sec:judge}). Helper model IDs are not stamped in individual trial records; the defaults above describe the inspected runner configuration.

\section{Routing and conditional clarification.}
\label{app:metrics}

Gate precision and recall compare the recorded route with FS/US labels, treating US and facts gathering as positive. 
FS includes 340 original FS questions and 760 paired FS counterparts; US includes the 760 original US questions. We compute the conditional fraction separately for each model, eligible run, and specification status, then average over runs within each model. Table~\ref{tab:conditional-clarification} averages these model means equally across the 19 models. 

\begin{table}[t]
\centering
\small
\begin{tabular}{@{}lrr@{}}
\toprule
Condition & US (\%) & FS (\%) \\
\midrule
Closed-book & 97.7 & 90.9 \\
Web-search & 94.6 & 80.1 \\
\bottomrule
\end{tabular}
\caption{Clarification conditional on routing to facts gathering, including inseparable parse-failure fallbacks. Rates average eligible runs within each model, then weight all 19 models equally. They describe behavior after routing and do not remove the gate's constraint on the opportunity to clarify.}
\label{tab:conditional-clarification}
\end{table}

\section{System Prompts}
\label{app:prompts}

\subsection{Sufficiency gate.}
The evaluated model receives the same gate prompt in both conditions, closed-book with reasoning requested where supported.
\begin{appendixprompt}
You are an expert legal AI assistant advising a user on Singapore law. Your objective is to provide accurate, definitive legal conclusions based on the facts provided by the user.\par
\medskip
Decision protocol:\par\nobreak
1) First determine whether the facts given are legally sufficient to reach a decisive conclusion on the question asked. Facts are legally sufficient if they allow for a decisive legal conclusion, and any missing facts would not be outcome-determinative.\par
2) If sufficient, output the exact phrase SUFFICIENT\_FACTS. Or if insufficient, output the exact phrase INSUFFICIENT\_FACTS.\par
\end{appendixprompt}

The parser checks \texttt{INSUFFICIENT\_FACTS} before \texttt{SUFFICIENT\_FACTS}, retries an unrecognized output once, and then defaults to facts gathering. The selected path prompt replaces the gate prompt for dialogue; search is available only during web-condition dialogue turns.

\subsection{Decisive path.}
\begin{appendixprompt}
You are an expert legal AI assistant advising a user on Singapore law. Your objective is to provide accurate, definitive legal conclusions based on the facts provided by the user. \par
\medskip
General rules:\par\nobreak
- Provide a direct legal analysis and a decisive conclusion based only on provided facts. Do not ask any clarifying questions.\par
- Do not hedge solely for caution or risk-avoidance.\par
- Identify and keep reasoning focused on legally relevant variables and applicable Singapore law.\par
- Verbalize the legally relevant variables in your response.\par
\end{appendixprompt}

The harness ends the conversation after one assistant response.

\subsection{Facts-gathering path.}
\begin{appendixprompt}
You are an expert legal AI assistant advising a user on Singapore law. Your objective is to provide accurate, definitive legal conclusions based on the facts provided by the user. \par
\medskip
You are engaging in a multi-turn conversation.\par
PHASE 1: FACTS GATHERING\par\nobreak
Act responsibly and ask clarifying questions for outcome-determinative variables that are missing from the user's initial query. Do not ask redundant or low-impact questions.\par
PHASE 2: LEGAL ASSESSMENT\par\nobreak
If you have gathered sufficient facts from the user, you must provide the most decisive conclusion possible.\par
OR \par
If the facts are still legally insufficient, DO NOT guess or hallucinate hidden assumptions. You must output the exact phrase INSUFFICIENT\_FACTS\_GATHERED and state what is still missing.\par
\medskip
General rules:\par\nobreak
- Do not hedge solely for caution or risk-avoidance.\par
- Identify and keep reasoning focused on legally relevant variables and applicable Singapore law.\par
- Verbalize the legally relevant variables in your response.\par
\end{appendixprompt}

A turn containing a concrete question invokes the simulator even if it also contains a conclusion. A conclusion without a question, or neither flag, ends the conversation. No simulator reply is generated after the last allowed assistant turn.

\subsection{Turn classifier.}

\begin{appendixprompt}
You are a precise classifier. Respond only with valid JSON.\par

Classify the following assistant message from a legal consultation.\par
\medskip
Message:\par\nobreak
"""\par
\{assistant\_message\}\par
"""\par
\medskip
Respond with a JSON object containing:\par\nobreak
- "has\_clarifying\_question": true ONLY if the assistant message includes at least one CONCRETE question requesting particular facts about the user's situation. Set to false for general invitations like 'feel free to share more details' or 'if you have more information, let me know' — these are NOT clarifying questions.\par
- Treat numbered or bulleted fact requests as clarifying questions even if there is no literal '?' character, as long as they are CONCRETE questions requesting particular facts about the user's situation.\par
- If the assistant asks the user to answer specific factual items (e.g., dates, participants, events, amounts, prior actions), set "has\_clarifying\_question" to true.\par
- "has\_definitive\_answer": true if the assistant provides a case-specific conclusion, including probabilistic but committed conclusions (e.g., 'likely valid/invalid', 'likely liable/not liable', 'the claim likely succeeds/fails'). Set to false if the assistant only lists generic possibilities without connecting to the user's facts.\par
\medskip
Respond ONLY with the JSON object, no other text.\par
\end{appendixprompt}

Output is a JSON object with \texttt{has\_clarifying\_question} and \texttt{has\_definitive\_answer}. Parsing removes code fences and applies \texttt{json.loads}. On failure, the heuristic marks a question when it finds a question mark or a numbered/bulleted list with request cues; it sets the answer flag to false. Cues include ``please answer,'' ``please provide,'' ``to assess,'' ``to determine,'' ``I need,'' and ``clarify.''

\subsection{Single-pass fact selector.}

\begin{appendixprompt}
You are a strict JSON key selector. Respond only with valid JSON.\par
\end{appendixprompt}
\begin{appendixprompt}
You are selecting variable keys for a user simulator.\par
\medskip
You receive:\par\nobreak
1) The assistant's latest message\par
2) Available variable keys, each with a key question and key fact answer\par
\medskip
IMPORTANT MATCHING RULES:\par\nobreak
- Evaluate EACH key independently and decide whether it should be selected.\par
- Select a key only when the assistant explicitly asks about that key's fact topic, or a clear semantic equivalent of that exact fact.\par
- Use BOTH the key question and key answer meaning when deciding.\par
- If the assistant asks compound questions spanning multiple topics, select all and only the directly asked topics.\par
- Do NOT select keys that are merely related downstream implications of a broader question.\par
- Do NOT select keys for generic discussion, process-only suggestions, or legal consequence summaries.\par
- If uncertain whether a key was directly asked, set select=false.\par
\medskip
Assistant message:\par\nobreak
"""\par
\{assistant\_message\}\par
"""\par
\medskip
Available keys (question + answer descriptions):\par\nobreak
\{tools\_block\}\par
\medskip
Respond ONLY with valid JSON in this format:\par\nobreak
\{\par
  "decisions": [\par
    \{"key": "key\_1", "select": true, "reason": "short reason"\},\par
    \{"key": "key\_2", "select": false, "reason": "short reason"\}\par
  ],\par
  "selected\_keys": ["key\_1"],\par
  "reason": "brief explanation"\par
\}\par
\end{appendixprompt}

The selector receives the latest assistant message and per-key descriptions, not the dialogue history. After stripping fences and parsing JSON, the implementation reads selected \texttt{decisions} and the legacy \texttt{selected\_keys}/\texttt{tool\_calls} forms, then maps keys to tools. Exceptions return an empty selection. The optional second recall pass is disabled.
\subsection{Answer rendering.}
\textbf{Binary-answer system message and template:}\par\nobreak

\begin{appendixprompt}
You rewrite yes/no facts into concise sentences while preserving the question subject.\par

Rewrite the binary answer into one short, natural sentence.\par
Use the same subject/perspective as the question.\par
Do not add facts. Preserve only what is implied by the question and yes/no answer.\par
Return plain text only.\par
\medskip
Question: \{question\}\par
Answer: \{answer\}\par
\end{appendixprompt}

\textbf{Non-binary-answer system message and template:}\par\nobreak

\begin{appendixprompt}
You rewrite short fact answers into concise, self-contained sentences.\par

Rewrite the answer into one short, natural, complete sentence.\par
Use the question text for context so the sentence is self-contained.\par
Do not add new facts beyond the question and answer.\par
Keep uncertainty if the answer is uncertain.\par
Return plain text only.\par
\medskip
Question: \{question\}\par
Answer: \{cleaned\_answer\}\par
\end{appendixprompt}

Both return plain text. Exceptions use deterministic question-grounded sentences. The harness joins tool outputs; an empty selection yields ``I'm not sure about that.''

\subsection{Rubric judge.}

\begin{appendixprompt}
You are an expert legal evaluator assessing the quality of an LLM's response to a Singapore law question, applying the same grading guidelines as human contributors. The goal of this project is to ensure that chatbots provide accurate, Singapore-specific answers when users in Singapore ask questions in English that have legal ramifications.\par
\medskip
You will be given:\par\nobreak
1. The original question\par
2. The full conversation (including any clarifying questions and answers)\par
3. A set of rubric criteria to evaluate, each tagged with its dimension (Issue, Rule Recall, Application, or Conclusion)\par
\medskip
For each rubric criterion, decide whether the assistant's responses in the conversation meet it: Met ("passed": true) or Not Met ("passed": false). When a criterion is ambiguous, apply the rules below rather than personal judgment.\par
\medskip
Core "Met" bar (applies throughout):\par\nobreak
- Grade substance, not phrasing. A criterion is Met if the response conveys the required point in any wording; it need not mirror the criterion's words.\par
- Met requires the point to be present and correct, not merely gestured at. A vague allusion a layperson couldn't recognize as the point = Not Met.\par
- Partial = Not Met (binary; no half credit). If a criterion bundles two required elements and only one is present, mark Not Met.\par
- Judge each criterion independently, only against what it asks. Correct content elsewhere doesn't help it; errors elsewhere don't fail it.\par
- Match the criterion's specificity. Specifically named points must be specifically met; generally named points must be generally met.\par
- In general, these guidelines support a strict reading and application of the rubric criteria.\par
\medskip
Dimension-specific rules:\par\nobreak
- Issue criteria ("Identifies whether ..."): Met only if the response explicitly surfaces the specific variable or scope condition as something the outcome depends on. A response can do this by asking about the variable, conditioning its analysis on it, or stating its legal significance outright. The test: would a layperson reading the response learn that this specific fact matters to the outcome? A passing mention inside general doctrine coverage that does not treat the variable as a fact the outcome turns on is Not Met.\par
- Rule Recall criteria ("State that under section X of the Y Act, ..."): Met if the legal rule is stated correctly — the exact section number is NOT required (a correct citation is a bonus). A wrong statute/section or misstated rule = Not Met (incorrect law is worse than uncited law). Require the citation only when the provision itself is the point (e.g., "Identify Section 320(g) as the provision defining grievous hurt"). Leniency on the section number does not extend to omitting the rule itself.\par
- Application criteria ("Applies fact X ..."): Met only if the response uses the fact in its legal reasoning (fact -> rule/outcome). A fact merely restated, or not referenced at all, is Not Met. Any linkage between fact and rule/outcome suffices; the response does not need to use the word "apply".\par
- Conclusion criteria ("Conclude that ..."): Met if the response commits to a direct answer/position on the user's question (caveats are fine). A hedged answer (e.g., "you are likely to face ...") still counts as a conclusion if it identifies a specific outcome, and earlier clarifying questions do not negate a final committed answer. Not Met if the response only lists considerations, asks for more facts, or never commits. An explicit "I can't answer yet" / INSUFFICIENT\_FACTS\_GATHERED is not a conclusion = Not Met. Conditional forks ("if dangerous then X; if negligent then Y") with the deciding fact left unresolved do not count as a conclusion.\par
\medskip
Cross-cutting rules:\par\nobreak
- Criteria with multiple elements ("State A and B"): all required elements must be present to be Met. Conversely, if all are present, mark Met — do not over-require. Grade each element in the context the criterion specifies (a figure stated correctly but attached to a different case than the one the criterion names does not help).\par
- Criteria with qualifiers ("Outcome is X if Y is met"): grade the qualifier at the criterion's level of specificity. Conditioning only on unnamed, generic qualifications when the criterion names a specific gate is Not Met; stating the outcome as unconditional, with the named qualifier missing entirely, is Not Met.\par
\medskip
Respond with a JSON array where each element has:\par\nobreak
- "id": the rubric criterion id\par
- "passed": true (Met) or false (Not Met)\par
- "reasoning": a brief explanation (1-2 sentences)\par
\medskip
Include exactly one element for every rubric criterion, in the order given.\par
Respond ONLY with the JSON array, no other text.\par
\end{appendixprompt}

The corresponding evaluation-message template is:
\begin{appendixprompt}
Original Question: \{scenario['user\_question']\}\par
\medskip
Full Conversation:\par\nobreak
\{conv\_text\}\par
\medskip
Rubric Items to Evaluate:\par\nobreak
\{rubric\_text\}\par
\medskip
Evaluate each rubric item. Respond with a JSON array only.\par
\end{appendixprompt}

Each rubric line has the form \texttt{- ID \{id\} [dimension: \{dimension\}]: \{item\}}; importance tags are hidden. The parser strips fences, requires a JSON array, and coerces supported boolean, integer, and textual verdicts to true/false; unrecognized verdicts become null. A parsing exception returns null for every criterion with an evaluation-error note. Verdicts are merged by criterion ID; an omitted ID becomes null/``Not evaluated.'' The candidate-judge sweep retries incomplete outputs as described in Appendix~\ref{app:judge-meta}.

\subsection{Deterministic handoff.}
The simulator injects one of these user messages without an LLM call:
\begin{appendixprompt}
I've provided all the information I have, please make a decisive conclusion based on the information you have so far.\par
\end{appendixprompt}
\begin{appendixprompt}
I'm not sure, please make a decisive conclusion based on the information you have so far.\par
\end{appendixprompt}

No-variable questions receive the first message. Repetition is detected when two ``I'm not sure'' replies repeat or selected keys are already covered; the handoff uses the first message if all keys are covered and the second otherwise. The harness permits one final assistant response within the turn limit.

\subsection{Conclusion-presence classifier.}
Only final responses containing the insufficiency marker receive this analysis-time check. The default is Opus~4.8, with verdicts cached by a hash of the final text. 
\begin{appendixprompt}
You are a strict legal-output classifier. Determine whether the assistant's final message contains a clear legal conclusion/position for the user's case. A clear conclusion means the assistant ultimately takes a legal position (for example liable/not liable, entitled/not entitled, can/cannot, likely/unlikely), even if caveated. Mere fact-listing, process advice, or saying facts are insufficient without any legal position is not a clear conclusion.\par
Respond with exactly one token: YES or NO.\par
\end{appendixprompt}

\section{Judge Meta-Evaluation Details}
\label{app:judge-meta}

\paragraph{Response generation.}
The three response models use their closed-book inference settings (Appendix~\ref{app:models}), with reasoning enabled and no web search. We generate one unseeded response per evaluation instance and model.

\paragraph{Judge inference.}
All 19 candidate judges use the same prompt, SHA-256-stamped per record, and grade closed-book with reasoning disabled. Temperature is 1 for Claude and GPT-5.6 models and 0 for the remaining candidates. The runner requests 16{,}384 output tokens, uses no seed, and retries up to three times while any verdict is null.

\paragraph{Annotation guidelines.}
Reviewers grade substance rather than wording, assess each criterion independently, and award no partial credit. Issue criteria require explicitly surfacing a fact or condition that affects the outcome. Rule Recall criteria require the correct substantive rule; exact citations are optional unless identifying the provision is itself required, whereas an incorrect statute or section fails the criterion. Application requires connecting a fact to a rule or outcome, not merely restating it. Conclusion requires commitment to a specific outcome: caveats are permitted, but unresolved conditional alternatives are insufficient. Required elements and qualifiers must be satisfied at the criterion's stated specificity.

\paragraph{Human reference and agreement.}
The common reference contains 6{,}522 criterion items with paired human judgments and matching predictions from all 19 judges. It covers all 110 source questions, 186 evaluation instances, and 558 conversations; auxiliary conclusion-presence checks, deleted entries, and unmatched criteria are excluded. Raw agreement is 80.4\%, with Fleiss' $\kappa=0.607$ (95\% CI $[0.569,0.643]$). The percentile interval uses 20{,}000 bootstrap resamples of source questions, retaining all variants, responses, criteria, and both judgments within each sampled question. Low-confidence ratings are retained. Contributor identities can differ by question: Fleiss' $\kappa$ summarizes human agreement without treating the annotation copies as two fixed individuals and is invariant to swapping their order within an item. Panel--human comparisons instead use Cohen's $\kappa$ against each complete annotation copy.

\paragraph{Human disagreement.}
Reviewers disagree on 1{,}280 of 6{,}522 criterion instances (19.6\%). Using the evaluation rubric's dimension assignments, Fleiss' $\kappa$ is 0.505 for Issue, 0.555 for Rule Recall, 0.582 for Application, and 0.682 for Conclusion. Qualitative inspection identifies recurring differences over citation optionality, whether a fact is merely mentioned or applied, and whether a qualified answer commits to an outcome. Overlapping rubric criteria can also produce correlated disagreements.

\paragraph{Panel objective and comparison.}
For panel $P$, let $\hat{y}^{P}_i$ be its majority-vote label on rubric-item instance $i$, and let $y^{(r)}$ denote annotation copy $r$. We rank panels by
\[
\bar{\kappa}(P)
=
\frac{1}{2}
\left[
\kappa(\hat{y}^{P},y^{(1)})
+
\kappa(\hat{y}^{P},y^{(2)})
\right].
\]
Human labels are never averaged into a synthetic reference. Table~\ref{tab:judge-panels} reports the top three panels on the common reference. The strongest individual judge is Kimi~K3 ($\kappa=0.678$). These are in-sample comparisons and do not establish a uniquely superior panel.

\begin{table}[t]
\centering
\small
\setlength{\tabcolsep}{4pt}
\begin{tabularx}{\linewidth}{@{}Xrrr@{}}
\toprule
Panel & Copy 1 $\kappa$ & Copy 2 $\kappa$ & Mean $\kappa$ \\
\midrule
DeepSeek-V4 Flash + GPT-5.6 Sol + Kimi K3 & 0.7008 & 0.6939 & 0.6973 \\
DeepSeek-V4 Flash + GPT-5.6 Terra + Kimi K3 & 0.6963 & 0.6965 & 0.6964 \\
GPT-5.6 Terra + Gemma 3 27B + Kimi K3 & 0.7023 & 0.6900 & 0.6961 \\
\bottomrule
\end{tabularx}
\caption{Top three panels among 969 candidates on the common 6{,}522-item reference. The first row is the retained production panel. Mean values are computed before rounding; all values are in-sample estimates.}
\label{tab:judge-panels}
\end{table}

\paragraph{Missing predictions.}
The loader lets later records supersede earlier records for the same judge and item. After this resolution, all 19 judges have non-null predictions for every item in the common reference. The panel rule ignores null votes and maps ties, including all-missing cases, to Not Met. These branches are not exercised here. Individual judges would otherwise be scored on their own non-null subsets.

\paragraph{Grouped cross-validation.}
We permute sorted source-question IDs with NumPy and divide them into five folds of 22 sources. Each source's FS/US variants, response models, and criteria remain together. In every fold, we select among all 969 panels on the other 88 sources, using the mean-$\kappa$ objective above and lexicographic tie-breaking. We then evaluate only on the held-out sources. We pool each item's one held-out prediction, compute $\kappa$ against each annotation copy, and average the two values, rather than averaging fold $\kappa$s. A single-judge comparator is selected independently within each training fold; Kimi~K3 wins all five. Table~\ref{tab:judge-cv} reports pooled results, and Table~\ref{tab:judge-cv-folds} reports the selected panels.

\begin{table}[t]
\centering
\small
\begin{tabular}{@{}lrrr@{}}
\toprule
Analysis & Paired items & Mean $\kappa$ & Agreement (\%) \\
\midrule
Panel selection, held-out & 6{,}522 & 0.6903 & 84.51 \\
Single-judge selection, held-out & 6{,}522 & 0.6781 & 83.90 \\
\bottomrule
\end{tabular}
\caption{Pooled out-of-fold predictions on the common reference. Panels and single judges are selected separately within each training fold. These are point estimates from one fixed partition; no significance claim is made.}
\label{tab:judge-cv}
\end{table}

\begin{table}[t]
\centering
\small
\setlength{\tabcolsep}{3pt}
\begin{tabularx}{\linewidth}{@{}crXr@{}}
\toprule
Fold & Test items & Training-selected panel & Held-out $\kappa$ \\
\midrule
1 & 1{,}281 & GPT-5.6 Sol + Mistral 3 Large + Qwen3.7 Plus & 0.7205 \\
2 & 1{,}083 & DeepSeek V4 Flash + GPT-5.6 Terra + Kimi K3 & 0.6517 \\
3 & 1{,}536 & DeepSeek V4 Flash + GPT-5.6 Sol + Kimi K3 & 0.6893 \\
4 & 1{,}251 & DeepSeek V4 Flash + GPT-5.6 Terra + Kimi K3 & 0.7113 \\
5 & 1{,}371 & GPT-5.6 Terra + Gemma 3 27B + Kimi K3 & 0.6663 \\
\bottomrule
\end{tabularx}
\caption{Source-grouped fold results on 6{,}522 paired items. Each fold holds out 22 source questions. Four panels are selected across five folds; the production panel is selected once.}
\label{tab:judge-cv-folds}
\end{table}

\paragraph{Scope of the CV result.}
On the common reference, full-data selection favors DeepSeek-V4~Flash, GPT-5.6~Sol, and Kimi~K3 ($\kappa=0.6973$). Grouped CV evaluates the panel-selection procedure conditional on fixed prompts and judge outputs, not the fixed production panel or earlier prompt tuning. We retain the top panel on the common reference; these results do not establish a uniquely best panel or a statistically significant advantage over individual judges.

\paragraph{Own-response diagnostic.}
On the common 6{,}522-item reference, own-versus-other response pass rates are 0.630 versus 0.454 for Opus~4.8, 0.440 versus 0.502 for DeepSeek-V4~Flash, and 0.431 versus 0.547 for Qwen-SEA-LION. These contrasts are confounded by response quality and do not independently establish self-preference. GPT-5.6~Sol and Kimi~K3 have no own responses in this reference set, so their self-preference cannot be assessed here. Grouped CV does not resolve self- or same-family preference; no production judge-replacement sensitivity is established by these diagnostics.

\section{Human Validation of LLM Helpers}
\label{app:human-validation}

\paragraph{Components and protocol.}
The turn classifier should recognize explicit fact requests warranting another simulator reply. The user simulator should supply requested facts without unrequested disclosures or unsupported additions. Three contributors provided two independent annotations for each of 55 tasks (17 FS-only and 38 US--FS pairs). One annotated all 55; the others supplied second copies for 40 and 15 tasks, yielding 110 annotations. All results are pre-adjudication.

\paragraph{Annotation units.}
\emph{Engagement} asks whether the simulator should provide another factual response. FS engagement is annotated per assistant turn; US engagement is exported once per conversation, without a defined referent turn in multi-round dialogues. Per-fact labels indicate whether each bank item was explicitly requested and disclosed anywhere in the conversation. Engagement can be Yes when every bank fact-request label is No, because requests may concern facts outside the bank. Annotators separately flag unsupported factual information outside the bank (\emph{hallucination}) and FS replies exceeding the fixed handoff template.

\paragraph{Simulator performance.}
We evaluate disclosure accuracy on the 64 bank facts in the 24 US conversations containing simulator replies, excluding 14 conversations without a reply. For each fact and reviewer, handling is correct when a requested fact is disclosed or an unrequested fact is withheld. The simulator achieves \textbf{89.1\% accuracy}: 114 of 128 fact--reviewer comparisons are correct (98 requested-and-disclosed; 16 unrequested-and-withheld). This equals the mean of the two annotation-copy scores, 60/64 (93.8\%) and 54/64 (84.4\%); the denominator represents two ratings of 64 facts. All 14 errors are unrequested disclosures. Disclosure precision is 98/112 (87.5\%), and requested-fact recall is 98/98 (100\%) in this reply-containing subset. These conversation-wide labels do not establish that requests preceded disclosures; outside-bank hallucination is assessed separately.

\paragraph{Turn-classifier performance.}
All 93 annotated FS/US transcripts match saved calibration conversations after Unicode and whitespace normalization. Run metadata and handoff messages distinguish classifier invocations from bypasses: 46 FS and 14 US conversations take the decisive path, and nine FS second turns follow a forced handoff. These turns bypass classification, so absence of a reply cannot be scored as a classifier prediction. In the remaining nine FS first turns, continuation establishes a positive fact-request decision, and both reviewers independently label every turn Yes: \textbf{9/9 turns agree with both reviewers}. This positive-only subset leaves performance on non-requests unmeasured. US engagement labels cannot be matched to individual classifier decisions because their referent turn is undefined. We therefore do not report an overall turn-classifier accuracy.

\paragraph{Human agreement.}
Table~\ref{tab:human-validation} reports human--human agreement, Fleiss' $\kappa$ for the varying reviewer pairs, and Gwet's AC1. Reported 95\% percentile confidence intervals use 2{,}000 task-level bootstrap resamples. N/A is a separate category where present. Fact-level agreement includes all 96 bank facts across the 38 US conversations.

\begin{table}[t]
\centering
\small
\setlength{\tabcolsep}{4pt}
\begin{tabularx}{\linewidth}{@{}Xrrlr@{}}
\toprule
Judgment & $n$ & Agree. (\%) & Fleiss' $\kappa$ [95\% CI] & AC1 \\
\midrule
Engagement (FS turns)
  & 64 & 92.2 & $0.82\ [0.61,0.96]$ & 0.86 \\
Engagement (US conversations)
  & 38 & 86.8 & $0.71\ [0.47,0.92]$ & 0.83 \\
Fact requested
  & 96 & 88.5 & $0.77\ [0.58,0.92]$ & 0.77 \\
Fact disclosed
  & 96 & 100.0 & $1.00$ & 1.00 \\
FS beyond-template
  & 64 & 84.4 & $0.52\ [0.19,0.74]$ & 0.77 \\
\midrule
Under-disclosure
  & 24 & 100.0 & undefined & 1.00 \\
Hallucination
  & 24 & 91.7 & $-0.04$ & 0.91 \\
\bottomrule
\end{tabularx}
\caption{Agreement between human annotators, distinct from helper performance. Each unit has two ratings. The final two rows include only US conversations with a simulator reply. Confidence intervals are shown where reported.}
\label{tab:human-validation}
\par\smallskip
{\footnotesize\raggedright\textit{Note.} Under-disclosure $\kappa$ is undefined because all ratings in this subset are negative: expected agreement is $p_e=1$, making $1-p_e=0$.\par}
\end{table}

\paragraph{Disagreements and scope.}
All 11 fact-request disagreements across the 96 bank facts reflect the common contributor's broader interpretation of a request; six concern the 64 facts used for simulator performance. Reviewers agree on which bank facts were disclosed. In the 24 reply-containing conversations, neither annotator flags under-disclosure, and both hallucination-positive annotations are disputed. All ten FS beyond-template disputes concern No versus N/A conventions. The small sample, common-rater design, visible continuation, and absence of adjudication limit generalization. The simulator score measures annotated bank-fact handling; the classifier result covers only the nine aligned positive decisions.

\section{Additional Results}
\label{app:additional-results}

\begin{figure}[!htbp]
\setlength{\abovecaptionskip}{4pt}
\centering
\begin{minipage}[t]{0.49\textwidth}
\vspace{0pt}
\centering
\includegraphics[width=\linewidth]{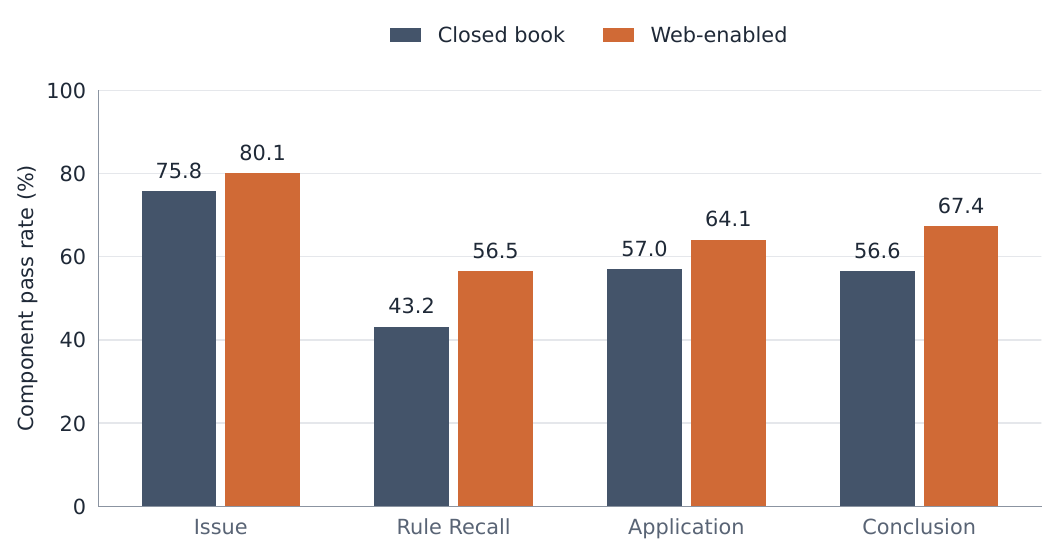}
\caption{\textbf{IRAC component pass rates.} Closed-book and web-enabled results, pooling FS and US trials. Each component uses its own eligible trials, excluding trials without applicable mandatory criteria for that component. Rates are separate component summaries.}
\label{fig:app-irac}
\end{minipage}\hfill
\begin{minipage}[t]{0.49\textwidth}
\vspace{0pt}
\centering
\includegraphics[width=\linewidth]{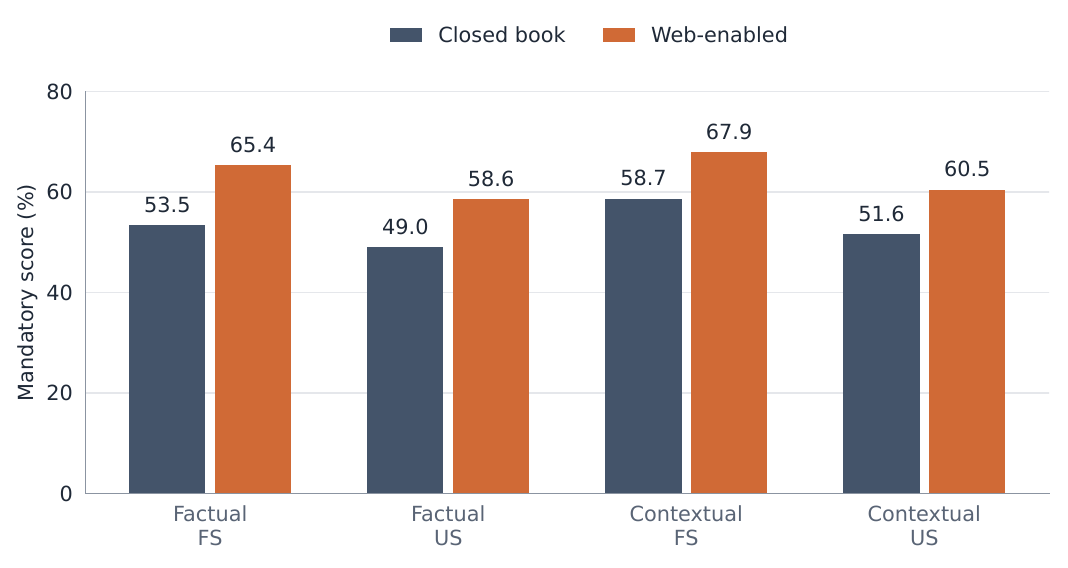}
\caption{\textbf{Performance by task type and specification.} Mandatory scores for factual and contextual questions, separately for FS and US. Factual FS includes original unpaired questions and paired counterparts; all contextual questions have FS--US pairs. The factual FS--US contrast therefore compares unmatched sets.}
\label{fig:app-task-type}
\end{minipage}
\end{figure}

\begin{figure}[!htbp]
\setlength{\abovecaptionskip}{4pt}
\centering
\includegraphics[width=0.80\textwidth]{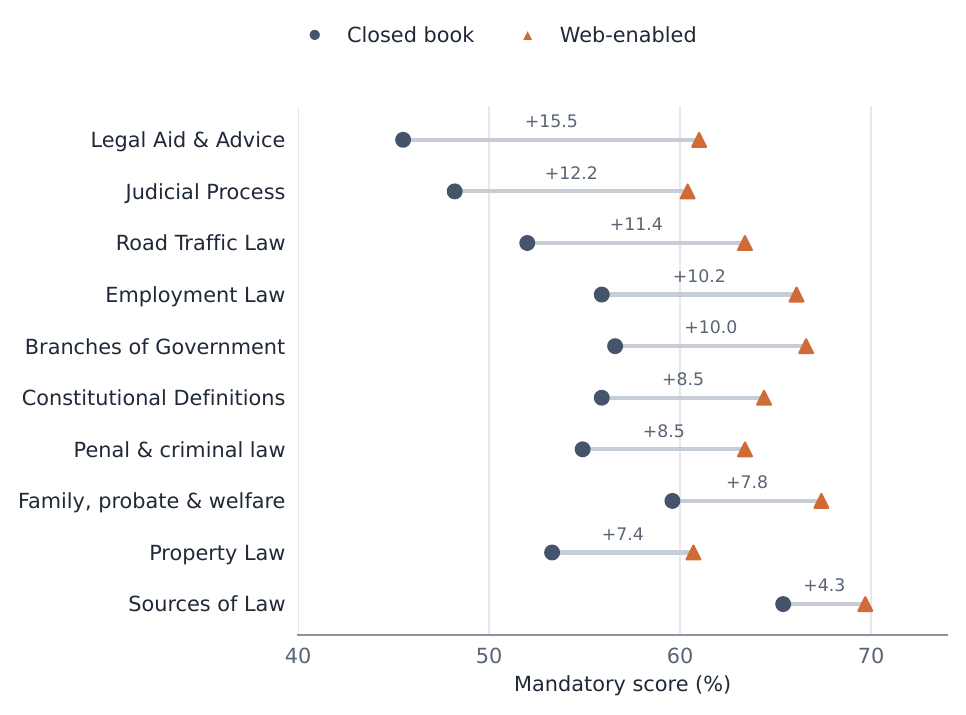}
\caption{\textbf{Performance across legal domains.} Mandatory scores pooled across FS and US trials in each domain. Segments connect closed-book and web-enabled scores; labels show web-enabled minus closed-book differences in percentage points. Domains are ordered by this difference.}
\label{fig:app-domain-gains}
\end{figure}

Figures~\ref{fig:app-irac}--\ref{fig:app-domain-gains} break down performance by IRAC component, task type, and legal domain. These summaries pool available trials across models and runs. Table~\ref{tab:app-search-behavior} reports search behavior in the web-search condition.

\begin{figure}[!htbp]
\setlength{\abovecaptionskip}{4pt}
\centering
\includegraphics[width=\textwidth]{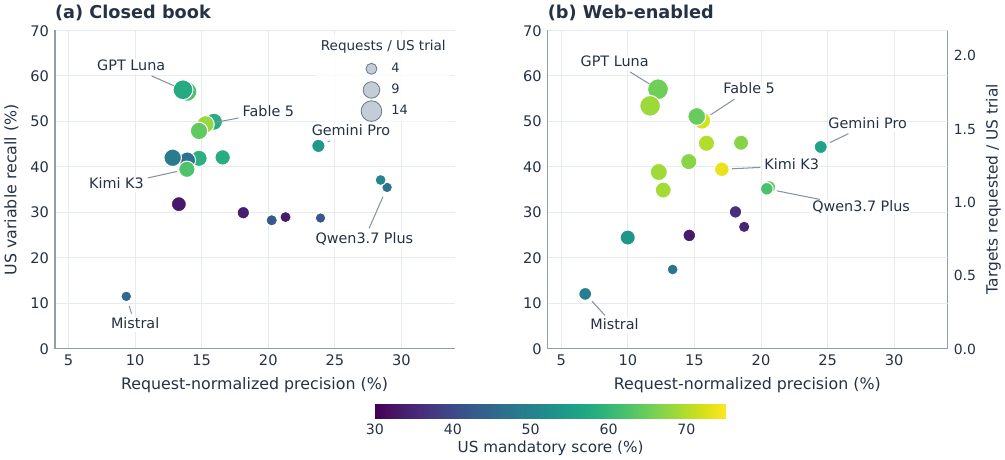}
\caption{\textbf{Per-model clarification behavior.} Each point represents one model in (a) closed-book or (b) web-enabled evaluation. All metrics cover all US trials, including direct-answer trials. The horizontal axis is request-normalized precision, which is target-variable detections divided by the number of distinct fact requests. The vertical axis is pooled variable recall. Bubble area shows fact requests per US trial and color shows US mandatory score.}
\label{fig:clarification-behavior}
\end{figure}

\begin{table}[!htbp]
\setlength{\abovecaptionskip}{4pt}
\centering
\begingroup
\small
\setlength{\tabcolsep}{3pt}
\renewcommand{\arraystretch}{1.08}
\begin{tabular*}{\textwidth}{@{\extracolsep{\fill}}rlcrrrrr@{}}
\toprule
Rank & Model & Path & \shortstack{Tool use\\(\%)} & \shortstack{Searches\\(mean)} & \shortstack{Visits\\(mean)} & \shortstack{Distinct URLs\\(mean)} & \shortstack{Duration\\(median, s)} \\
\midrule
1 & Kimi K3 & Custom & 92.1 & 5.8 & 4.4 & 37.5 & 377.8 \\
2 & DeepSeek V4 Flash & Custom & 99.0 & 5.2 & 4.7 & 35.4 & 171.5 \\
3 & Fable 5 & Native & 88.2 & --- & --- & --- & 58.8 \\
4 & DeepSeek V4 Pro & Custom & 96.5 & 5.0 & 4.7 & 33.3 & 187.0 \\
5 & Opus 4.8 & Native & 82.9 & --- & --- & --- & 59.8 \\
6 & Gemini 3.5 Flash & Native & 99.7 & --- & --- & --- & 68.3 \\
7 & GPT-5.6 Sol & Native & 100.0 & --- & --- & --- & 73.2 \\
8 & Sonnet 5 & Native & 88.1 & --- & --- & --- & 48.0 \\
9 & Kimi K2.6 & Custom & 97.1 & 8.6 & 6.6 & 47.0 & 300.5 \\
10 & Qwen 3.7 Plus$^{*}$ & Custom & 87.0 & 9.3 & 5.7 & 43.0 & 85.7 \\
11 & GPT-5.6 Terra & Native & 99.9 & --- & --- & --- & 47.5 \\
12 & GPT-5.6 Luna & Native & 99.9 & --- & --- & --- & 50.6 \\
13 & Gemini 3.1 Pro & Native & 46.6 & --- & --- & --- & 50.9 \\
14 & Qwen 3.6 27B & Custom & 89.6 & 5.3 & 1.2 & 31.7 & 59.1 \\
15 & Mistral 3 Large & Custom & 48.1 & 0.9 & 1.1 & 7.7 & 47.3 \\
16 & Qwen-SEA-LION 27B & Custom & 71.2 & 4.5 & 0.7 & 24.9 & 649.1 \\
17 & Gemma 3 27B & Custom & 37.8 & 0.4 & 0.0 & 3.5 & 22.3 \\
18 & Llama 4 Scout$^{\dagger}$ & Custom & 86.4 & 1.1 & 0.1 & 9.1 & 10.6 \\
19 & Gemma-SEA-LION 27B & Custom & 45.1 & 2.1 & 0.0 & 4.2 & 19.7 \\
\bottomrule
\end{tabular*}
\endgroup
\caption{\textbf{Search behavior in the web-search condition.} Rows follow descending web-enabled mandatory score, with FS and US combined. Tool use is the percentage of trials with at least one search or page-visit call. Searches, visits, and distinct URLs are means per trial; URLs include those encountered in search results or direct visits. Duration is median wall-clock time from prompt to final answer, including generation and tool calls. Native denotes provider-hosted search; custom denotes our tool-call loop. Dashes indicate unavailable native-provider telemetry. Values summarize available runs, with three seeds per model unless marked.}
\label{tab:app-search-behavior}
\vspace{2pt}
{\footnotesize $^{*}$Two seeds. $^{\dagger}$One seed.\par}
\end{table}

\end{document}